\documentclass{article}

\usepackage[preprint]{neurips_2019}

\usepackage[utf8]{inputenc} 
\usepackage[T1]{fontenc}    
\usepackage{hyperref}       
\usepackage{url}            
\usepackage{booktabs}       
\usepackage{amsfonts}       
\usepackage{nicefrac}       
\usepackage{microtype}      
\usepackage{hyperref}
\usepackage{url}
\usepackage{amssymb} 
\usepackage{amsthm} 
\usepackage{amsmath}
\usepackage{epstopdf} 
\usepackage{float}
\usepackage{mathtools}
\usepackage{natbib}
\usepackage{xcolor}
\usepackage{subcaption}

\DeclareMathOperator*{\E}{\mathbb{E}}

\title{Contrastive World Models}

\author{%
  Bonnie Li \\
}

\begin{document}

\maketitle

\begin{abstract}
World models trained via pixel reconstruction can struggle in visually complex environments, where irrelevant information dominates the objective and distract the model from information relevant to planning and control. We present Contrastive World Models, an approach for learning latent dynamics models without pixel reconstruction. Building on Dreamer, we replace observation reconstruction in the standard world model objective with a Deep InfoMax-like lower bound that maximizes the mutual information between state-action sequences and local patch features of future observations, encouraging state representations to retain information that is predictive of the future without requiring the model to reconstruct visually irrelevant details. We evaluate our approach in small-scale experiments across three settings of increasing visual complexity. Our method matches Dreamer and a momentum prediction baseline in the default setting, and substantially outperforms both once distractors or natural video backgrounds are introduced, while also training more efficiently by removing the pixel decoder entirely. Our approach is general and makes minimal assumptions beyond access to state-action sequences and future observations. These results suggest that contrastive, infomax-based objectives are a principled and promising direction for building world models that are robust to visual nuisance factors, a property particularly relevant for transferring model-based RL agents to the real world.
\end{abstract}

\begin{figure}[htbp]
\centering
\includegraphics[width=0.8\textwidth]{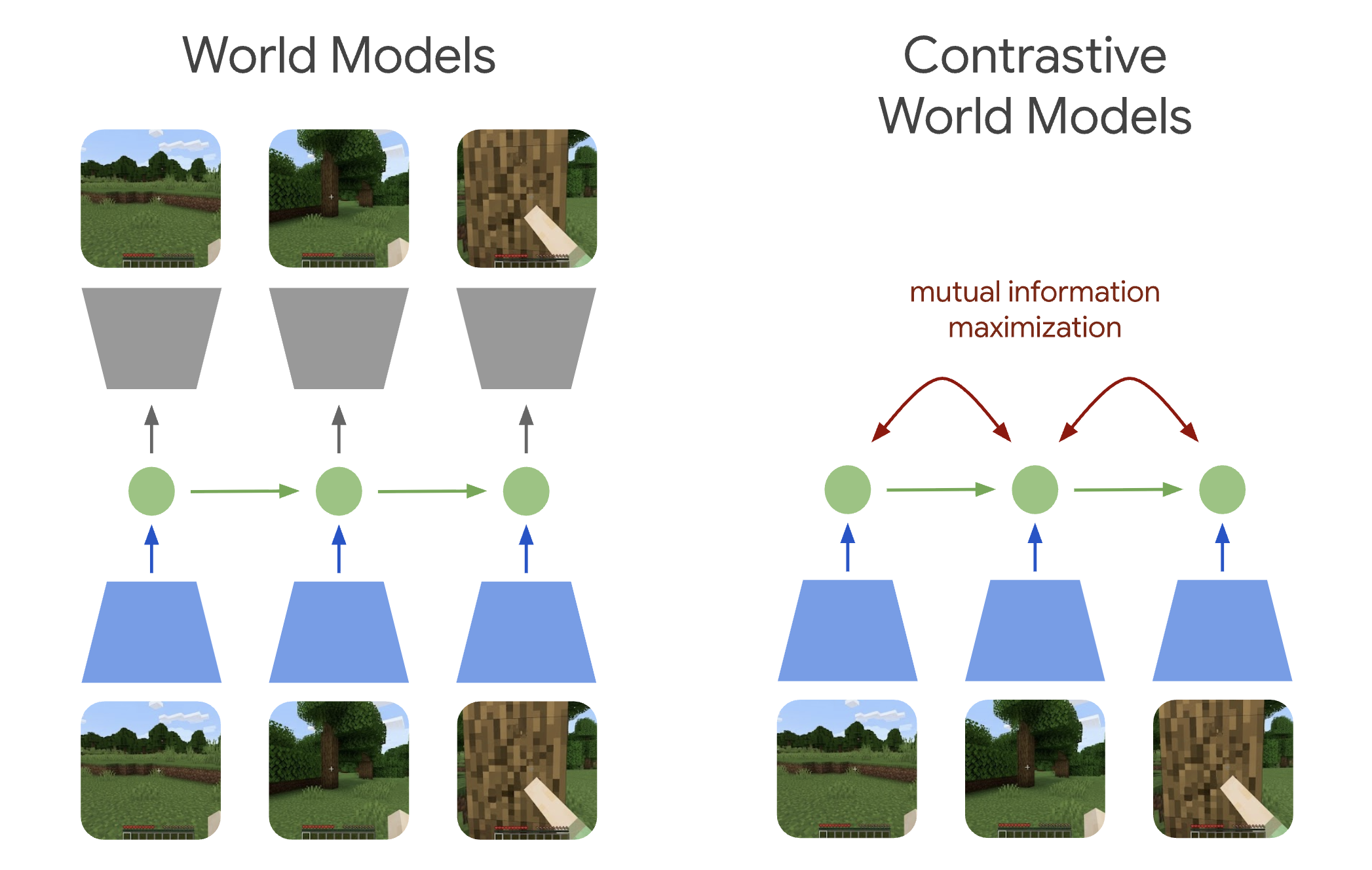}
\caption{Pixel-prediction-based World Models (left) and Contrastive World Models (right). Standard world models encode each observation into a latent state (green) via an encoder (blue) and reconstruct the input image through a pixel decoder (gray). Contrastive World Models remove the decoder entirely and instead train the same encoder and latent dynamics model by maximizing mutual information between latent states and local features of future observations.}
\label{fig:highlevel}
\end{figure}

\section{Introduction}
Learning a model of the environment's dynamics -- a \textit{world model} -- has emerged as a central paradigm for building agents that can plan, imagine, and act efficiently from high-dimensional observations. \citet{ha2018worldmodels} showed that an agent could be decomposed into a vision component (a variational autoencoder), a memory component (a recurrent network), and a small controller, and that the controller could be trained entirely "inside" imagined rollouts of the learned model before being transferred back to the real environment. World models have now become a practical recipe and a broader direction that has been adopted well beyond simple control tasks, such as for interactive world generation \citep{genie2, genie3}, autonomous driving \citep{zhang2024copilot, russell2025gaia2}, and robotics \citep{yang2024learninginteractiverealworldsimulators, ye2026worldactionmodelszeroshot}.

The Dreamer and Genie series has driven much of the progress along this direction. \citet{planet} introduced the Recurrent State Space Model (RSSM), combining deterministic and stochastic recurrent latents for planning via model-predictive control; \citet{dreamer} built on this by learning a value function and actor from imagined rollouts, replacing planning with amortized policy learning; and DreamerV2 \citep{dreamerv2} and DreamerV3 \citep{dreamerv3} progressively improved robustness on top of the RSSM backbone, with DreamerV3 mastering over 150 tasks under a single fixed hyperparameter set. In parallel, the Genie series has pushed toward large-scale, action-controllable video generation. Genie \citep{genie1} learns to generate interactive 2D environments purely from unlabeled video, Genie 2 \citep{genie2} extends this to diverse 3D environments, and Genie 3 \citep{genie3} generates real-time, minutes-long interactive worlds  from text alone. More recently, SIMA 2 \citep{sima2} demonstrated an embodied agent operating directly within Genie 3-generated worlds. All these works use  reconstruction or pixel-prediction objectives to learn latent state representations.

Contrastive learning offers a natural alternative to reconstruction. Rather than regenerating its input, contrastive objectives train an encoder to distinguish related ("positive") pairs of views or timesteps from unrelated ("negative") ones, typically by optimizing the InfoNCE lower bound on mutual information \citep{oord2019representation}. This principle underlies much of the progress in self-supervised visual representation learning: Contrastive Predictive Coding \citep{oord2019representation} and Deep InfoMax \citep{hjelm2019learning} maximize mutual information between global and local features of an input; SimCLR \citep{chen2020simclr} and MoCo \citep{he2020moco} scale instance discrimination with large or memory-bank negative sets; and BYOL \citep{grill2020byol} replaces explicit negatives with a momentum target network and prediction task. For world models, these objectives never require pixels to be reconstructed: the encoder only needs to preserve information that discriminates the correct future from alternatives. Similar motivations have driven representation learning in model-free RL, where DrQ \citep{yarats2021drq}, ATC \citep{stooke2021atc}, ST-DIM \citep{anand2020unsupervised}, DRIML \citep{mazoure2020deepreinforcementinfomaxlearning}, and MPR \citep{schwarzer2021mpr} use augmentation-based, contrastive, or momentum-based auxiliary objectives to improve sample efficiency, though without an explicit dynamics model for planning.

There has recently been a resurgence of interest in learning world models
directly in latent space, without pixel-level generation. Notably,
\citet{dreamer} evaluated a contrastive variant of Dreamer and found that it
underperformed pixel reconstruction on the majority of tasks at the time.
Another early work is CSWM \citep{kipf2020cswm}, which encodes each
observation into a set of object slots and models transitions with a graph
neural network over these slots, trained with a TransE-style energy-based
hinge loss with negative sampling; however, it is evaluated only on latent
ranking metrics in simple deterministic domains, without rewards or
downstream control. The Joint Embedding Predictive Architecture (JEPA)
\citep{lecun2022path} pursues the same principle at scale, predicting the
latent embedding of a target view from a context view so that unpredictable
low-level detail is discarded, with instantiations for images
\citep{assran2023ijepa} and video \citep{bardes2024vjepa, assran2025vjepa2}.
Closest to our setting, \citet{maes2026leworldmodel} introduce LeWorldModel
(LeWM), a JEPA-based world model that trains stably end-to-end from raw
pixels using only a next-embedding prediction loss and a Gaussian
regularizer, dispensing with the stop-gradients, momentum encoders, and
multi-term losses earlier variants required to prevent collapse. 
However, it has so far been validated only at small scale in visually clean domains---LeWM is a ${\sim}15$M-parameter model evaluated on four low-distraction environments with stationary backgrounds (2D maze navigation, Reacher, Push-T, and OGBench-Cube)---and has not been shown to be robust to complex visual nuisance factors, which our work addresses.

In this work, we combine these two directions -- contrastive representation learning and Dreamer world model -- by replacing the reconstruction-based observation model in Dreamer with a mutual information maximization objective similar to Deep InfoMax, removing the pixel decoder entirely. We hypothesize that this approach should have little effect in visually simple settings but should yield substantially more robust representations and more efficient training, and therefore stronger downstream control performance, as visual complexity and distraction increase. 


\section{Contrastive World Models}
We consider a partially observable Markov decision process (POMDP). We define a discrete time step t, image observations $o_t$, hidden states $s_t$, continuous action vectors $a_t$, and scalar rewards $r_t$.
We seek to build a latent dynamic model of the environment, of which an agent can use to plan and to maximize its expected cumulative reward. To do this, we build on top of a strong model-based RL algorithm, Dreamer \citep{dreamer}, and propose an alternative InfoMax objective to learn robust state representations. We now describe our approach in detail.

\subsection{Latent Dynamics Model}
We consider learning a latent dynamics model for a POMDP with image observations $o_t$,
continuous actions $a_t$, scalar rewards $r_t$, and latent states $s_t$. The generative
model, or ``world model,'' factorizes as
\begin{align*}
\text{transition model} &\quad s_t \sim p(s_t \mid s_{t-1}, a_{t-1}) \\
\text{observation model} &\quad o_t \sim p(o_t \mid s_t) \\
\text{reward model} &\quad r_t \sim p(r_t \mid s_t) \\
\text{state representation model (encoder)} &\quad s_t \sim q(s_t \mid s_{t-1}, a_{t-1}, o_t)
\end{align*}

We would like to fit this model by maximizing the marginal log-likelihood of the observed
data, $\ln p(o_{1:T}, r_{1:T} \mid a_{1:T})$, but this requires marginalizing out the full
latent trajectory $s_{1:T}$ and is intractable. We instead derive a tractable lower bound
by introducing the variational posterior above and applying Jensen's inequality.

First, we rewrite the marginal likelihood as an expectation under $q$ by multiplying and
dividing by $q(s_{1:T} \mid o_{1:T}, a_{1:T}) = \prod_t q(s_t \mid s_{t-1}, a_{t-1}, o_t)$:
$$
\ln p(o_{1:T}, r_{1:T} \mid a_{1:T})
= \ln \mathbb{E}_q \left[ \frac{p(o_{1:T}, r_{1:T}, s_{1:T} \mid a_{1:T})}{q(s_{1:T} \mid o_{1:T}, a_{1:T})} \right]
$$
Since $\ln(\cdot)$ is concave, Jensen's inequality gives $\ln \mathbb{E}[X] \geq \mathbb{E}[\ln X]$, so
$$
\geq \; \mathbb{E}_q \left[ \ln p(o_{1:T}, r_{1:T}, s_{1:T} \mid a_{1:T}) - \ln q(s_{1:T} \mid o_{1:T}, a_{1:T}) \right]
$$
Expanding both terms using the factorizations of the generative model and the posterior,
$$
= \; \mathbb{E}_q \left[ \sum_{t} \bigg(
\ln p(o_t \mid s_t) + \ln p(r_t \mid s_t) + \ln p(s_t \mid s_{t-1}, a_{t-1}) - \ln q(s_t \mid s_{t-1}, a_{t-1}, o_t)
\bigg) \right]
$$
and grouping the last two terms per timestep into a KL divergence yields our final bound,
$$
= \; \mathbb{E}_q \left[ \sum_{t} \bigg(
\ln p(o_t \mid s_t) + \ln p(r_t \mid s_t) -KL\Big( q(s_t \mid s_{t-1}, a_{t-1}, o_t) \parallel p(s_t \mid s_{t-1}, a_{t-1}) \Big)
\bigg) \right]
$$

\begin{figure}[ht]
\begin{center}
\includegraphics[width=0.52\textwidth]{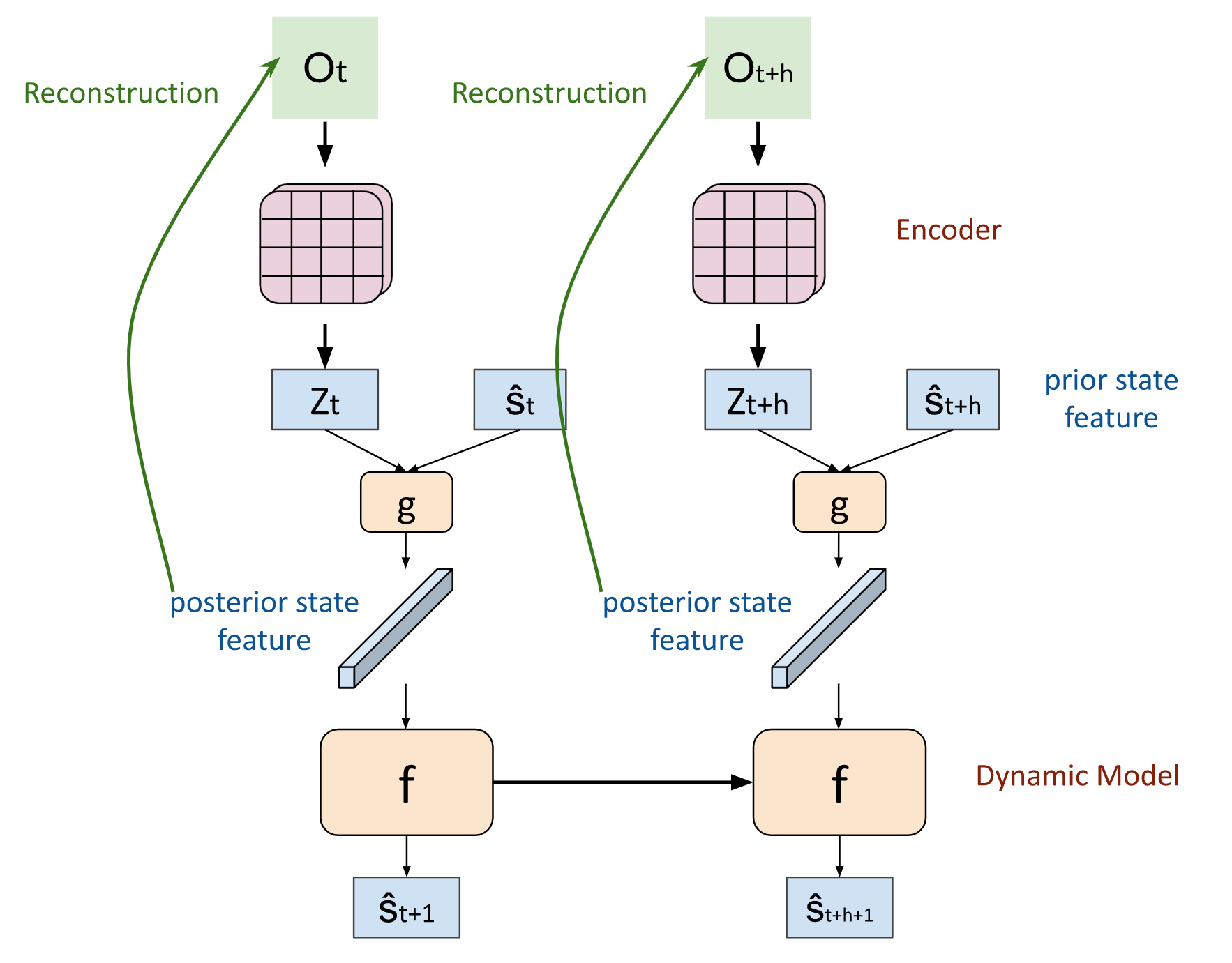}
\includegraphics[width=0.45\textwidth]{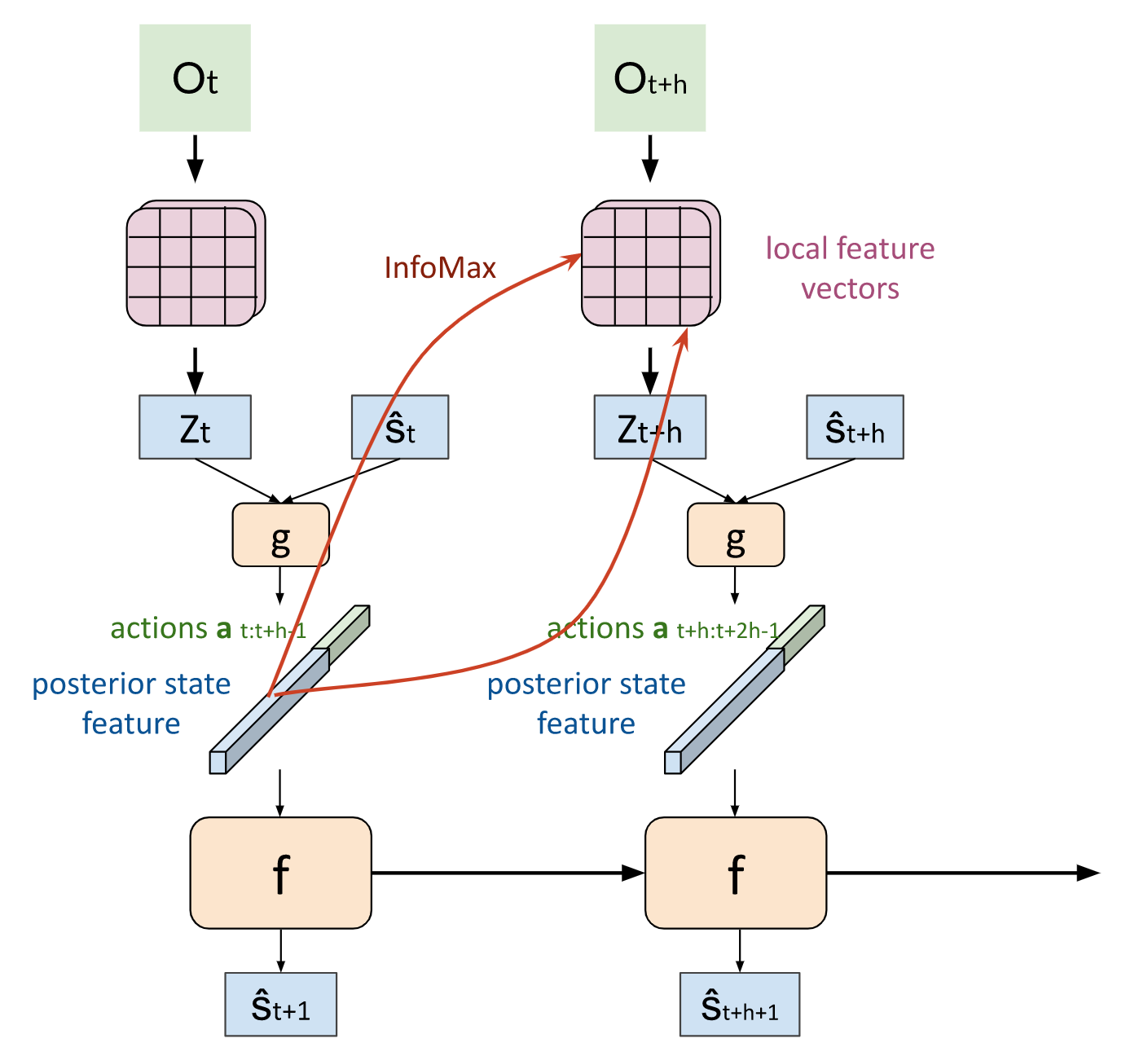}
\end{center}
\caption{Architecture of Recurrent State Space Model (left) and Contrastive World Model (right)}
\label{fig:architecture}
\end{figure}

Specifically, we build on top of Recurrent State Space Model \citep{planet}. We now describe the architecture in detail, as shown in figure \ref{fig:architecture}.

Given observation $o_t$ from timestep $t$, encoder $e$ encodes the high-dimensional observation to a lower dimension vector, denoted by $z_t$.
The update function $g$ takes encoding $z_t$ and the predicted prior state features and outputs the posterior state features, denoted $s_t$.
The transition function $f$ concatenates action $a_t$ and the posterior state features $s_t$ and predicts the prior state features for the next state, denoted as $\hat{s}_{t+1}$. The prior prediction is Gaussian with mean and variance parameterized by the model output.
Furthermore, the reward function $R$ maps the posterior state feature to rewards. The reward prediction is Gaussian with mean parameterized by the model output and unit variance.

For the observation likelihood term, RSSM uses a decoder $O$ that reconstructs the observation $o_t$ from the posterior state features $s_t$.

The training objective becomes observation reconstruction, reward prediction, and a KL regularizer. All the components of RSSM are optimized jointly.
\begin{center}
$J_{REC} = \mathbb{E}_q\left( \sum_{t} (J_O^t + J_R^t + J_D^t) \right)$
\end{center}
where
$$
J_O^t = \ln \left( O(o_t \mid s_t) \right)
$$
$$
J_R^t = \ln \left( R(r_t \mid s_t) \right)
$$
$$
J_D^t
= -KL\big( q(s_{t+1} \mid s_t, a_t, o_{t+1}) \parallel f(\hat{s}_{t+1} \mid s_t, a_t) \big)
$$

\subsection{InfoMax Representations}
Consider the observation likelihood term in the lower bound. We can subtract the observation marginal, which does not depend on $s_t$ and so does not shift the optimum with respect to the encoder, and apply Bayes' rule.
\begin{equation}
\begin{split}
\mathbb{E} \left[ \ln{ \left( p(o_t \mid s_t) \right)} \right]
& \doteq \mathbb{E} \left[ \ln{ p(o_t \mid s_t)} - \ln{ p(o_t)} \right] \\
& = \mathbb{E} \left[ \ln{ \frac{p(o_t, s_t)}{p(o_t)\, p(s_t)}} \right]
\end{split}
\end{equation}
Remark that the resulting density ratio is the mutual information between the observation $o_t$ and posterior state features $s_t$. This suggests mutual information maximization as an alternative objective for learning state representations in world models, without requiring an explicit pixel decoder.

To learn temporally dependent representations, we want to maximize mutual information between state-action sequences and future observations in the trajectory.
$$\max_\theta I([s_t, a_{t:t+h}]; o_{t+h})$$
which is bounded by the InfoNCE lower bound up to a constant
$$ \mathbb{E}\left[ \ln \frac{f_\theta([s_t, a_{t:t+h}], o_{t+h}) }
{\sum_{t'} f_\theta([s_t, a_{t:t+h}], o_{t'})}\right] $$
where $f_\theta$ is a score function which preserves mutual information, i.e.
$f_\theta(x_{t+h}, c_t) \propto \frac{p(x_{t+h}\mid c_t)}{p(x_{t+h})}$ when optimizing the above bound \citep{oord2019representation}. While any score function can be used, a simple log-bilinear model usually suffices:
$$f_\theta(x_{t+h}, c_t) = \exp(c_t^\top W_\theta x_{t+h})$$

Following Deep InfoMax \citep{hjelm2019learning}, in our setting we maximize mutual information between the global state posterior $s_t$ and local features of future states, i.e., patches of the encoding $e_{m,n}(o_{t+h})$. Here we concatenate the global state posterior $s_t$ with the action sequence $a_{t:t+h}$ leading up to the future state.

\begin{equation}
\max_\theta I([s_t, a_{t:t+h}]; o_{t+h})
\geq \mathbb{E}_{T, B} \left( \frac{1}{M} \frac{1}{N} \sum_m \sum_n
\ln \frac{\exp([s_t, a_{t:t+h}]^\top W_\theta e_{m, n}(o_{t+h}) )}
{\sum_{b \in B} \exp([s_t, a_{t:t+h}]^\top W_\theta e_{m, n}(o_b) )} \right)
\end{equation}

where $T, B$ denote the time and batch dimension of the sampled sequence respectively, and $m, n$ denote the $m$th and $n$th patch of the spatial features of the encoding.
For the training objective, we have
$$
J_I^t = \frac{1}{M} \frac{1}{N} \sum_m \sum_n
\ln \frac{\exp([s_t, a_{t:t+h}]^\top W_\theta e_{m, n}(o_{t+h}))}
{\sum_{b \in B} \exp([s_t, a_{t:t+h}]^\top W_\theta e_{m, n}(o_b) )}
$$
$$ J_R^t = \ln \left( R(r_t \mid s_t) \right) $$
$$ J_D^t = -KL\big( q(s_{t+1} \mid s_t, a_t, o_{t+1}) \parallel f(\hat{s}_{t+1} \mid s_t, a_t) \big) $$
$$J_{DIM} = \mathbb{E} \left(
\frac{1}{T} \sum_{t} (J_I^t + \lambda_1 J_D^t + \lambda_2 J_R^t) \right)$$

where $J_I^t$ is the Deep InfoMax representation loss, $J_R^t$ is the reward prediction objective, and $J_D^t$ is the KL transition objective. $\lambda_1, \lambda_2$ scale the transition loss and reward loss respectively.


\section{Experiments}
\begin{figure}[ht]
\begin{center}
\includegraphics[width=0.15\textwidth]{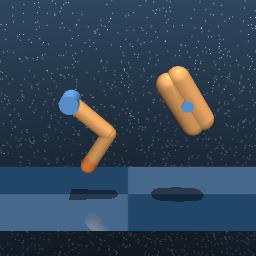}
\includegraphics[width=0.15\textwidth]{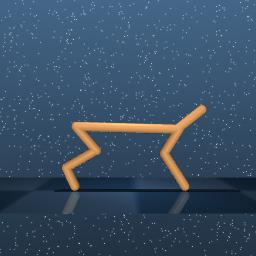}
\includegraphics[width=0.15\textwidth]{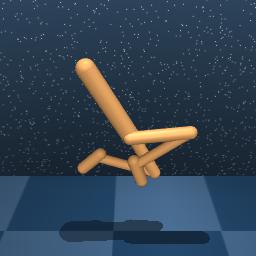}
\includegraphics[width=0.15\textwidth]{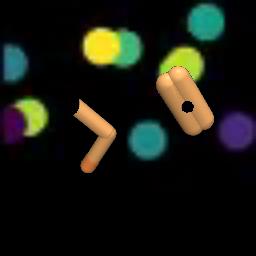}
\includegraphics[width=0.15\textwidth]{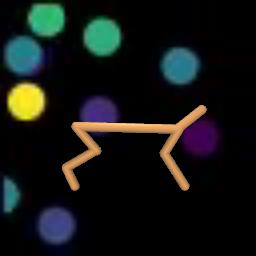}
\includegraphics[width=0.15\textwidth]{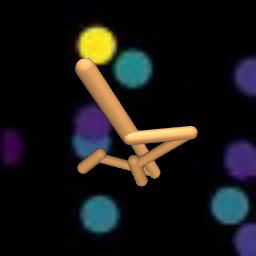}
\includegraphics[width=0.15\textwidth]{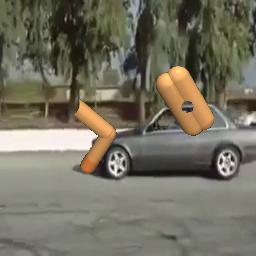}
\includegraphics[width=0.15\textwidth]{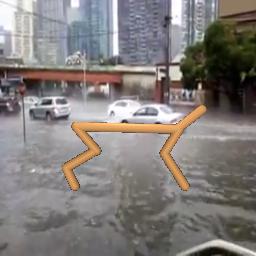}
\includegraphics[width=0.15\textwidth]{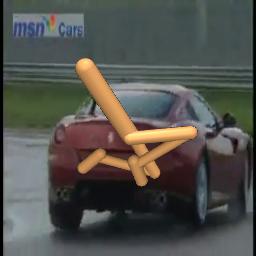}
\end{center}
\caption{Evaluation settings. Upper left three: default DMC setting; Upper right three: simple distractor setting; Lower three: natural video setting.}
\label{fig:dmc_settings}
\end{figure}

Our hypothesis is that our infomax approach should learn substantially more robust representations in visually diverse environments. To this end, we evaluate our approach in three settings shown in Figure \ref{fig:dmc_settings}, results are averaged over 3 seeds.
\textbf{Default Setting}: this is the default setting from DeepMind Control Suite with stationary background. 
\textbf{Simple Distractor Setting}: we incorporate simple moving distractors in the background, consisted of colored balls moving and bouncing off the frames \citep{yarats2020improving, zhang2021learning}. 
\textbf{Natural Video Setting}: We incorporate natural video from the Kinetics dataset as background, as done in \citep{zhang2021learning}. Note that in DBC \citep{zhang2021learning}, the natural background are grayscale, here we use full color channels without grayscale, which could be more challenging for the agents.

\begin{figure}[ht]
\includegraphics[width=1\textwidth]{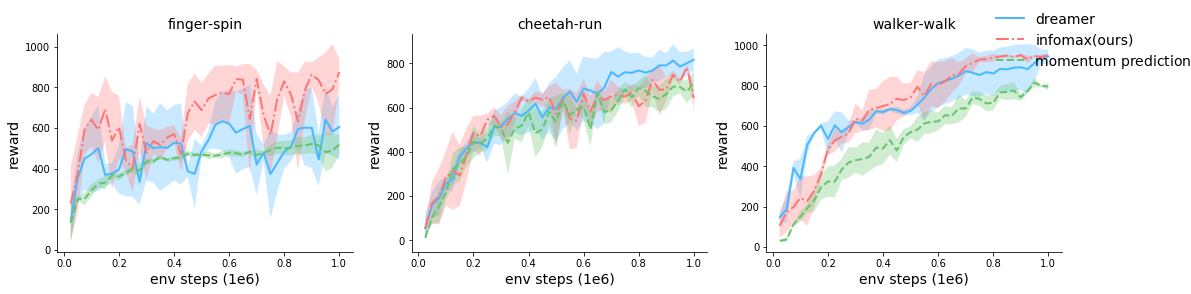}
\caption{\textbf{Default DMC setting}.}
\label{fig:default}
\includegraphics[width=1\textwidth]{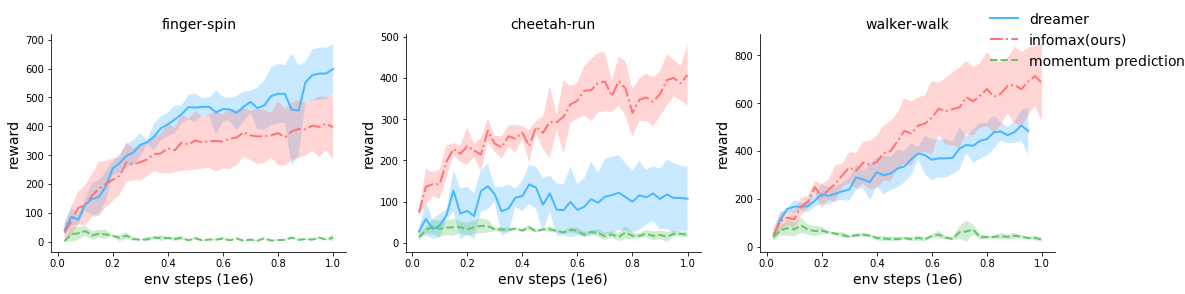}
\caption{\textbf{Simple distractor setting}.}
\label{fig:distractor}
\includegraphics[width=1\textwidth]{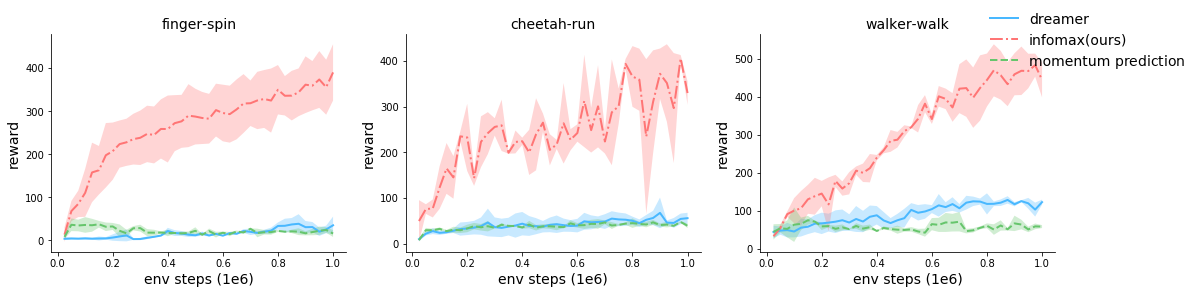}
\caption{\textbf{Natural video setting}.}
\label{fig:natural}
\end{figure}

\subsection{Baselines and Experimental Setup}
We compare our Contrastive World Model (\textbf{InfoMax}, ours) against two baselines: \textbf{Dreamer} \citep{dreamer}, which learns state representations via pixel reconstruction, and \textbf{Momentum Prediction}, a baseline that replaces the reconstruction objective with a momentum-encoder-based target but otherwise shares the same RSSM backbone and actor-critic as our method. 

\paragraph{Momentum Prediction baseline.}
As a second baseline, we consider a self-predictive representation objective similar to BYOL \citep{grill2020byol} and MPR \citep{schwarzer2021mpr}, which we refer to as Momentum Prediction. Rather than reconstructing pixels or maximizing an InfoNCE bound, this baseline maintains a momentum-averaged target encoder $e_\xi$, whose parameters $\xi$ are an exponential moving average of the online encoder $e$'s parameters $\theta$:
$$
\xi \leftarrow \tau \xi + (1 - \tau) \theta
$$
Given the RSSM state $s_t$ (the concatenation of the deterministic belief and stochastic posterior components), a projection head $\phi_\theta$ predicts the embedding of a future observation $o_{t+h}$, and the online prediction is trained to match the stop-gradient target embedding produced by the momentum encoder:
$$
J_{MP}^t = \cos\Big(\phi_\theta(s_t),\ \mathrm{sg}\big(e_\xi(o_{t+h})\big)\Big)
$$
where $\mathrm{sg}(\cdot)$ denotes the stop-gradient operator and $\cos(\cdot, \cdot)$ is cosine similarity. Unlike our InfoMax objective, $J_{MP}^t$ contains no explicit negative samples; representational collapse is instead avoided through the asymmetry between the online projector and the momentum target, following \citet{grill2020byol}. As with our method, this loss is combined with the same reward and transition losses used elsewhere:
$$
J_{MP} = \E \left( \frac{1}{T} \sum_t (J_{MP}^t + \lambda_1 J_D^t + \lambda_2 J_R^t) \right)
$$

All three agents use an identical actor-critic planner on top of their respective learned latent dynamics, and differ only in the world model representation-learning objective. Each agent is trained for $1\times10^{6}$ environment steps on three DeepMind Control Suite tasks (\texttt{finger-spin}, \texttt{cheetah-run}, \texttt{walker-walk}), and all results are averaged over 3 seeds.

\subsection{Results}

\textbf{Default DMC setting.} In the default, distraction-free setting (Figure~\ref{fig:default}), all three methods eventually reach comparable asymptotic performance, indicating that when the observation is uncluttered, pixel reconstruction provides a sufficient learning signal and our InfoMax objective provides near-equivalent performance. Momentum Prediction is competitive with the other two methods in this setting, confirming that BYOL objectives are viable state representation learners when the observation is simple.

\textbf{Simple distractor setting.} When simple moving distractors are added. (Figure~\ref{fig:distractor}), a clear separation emerges. InfoMax matches or exceeds Dreamer on two tasks, most notably on \texttt{cheetah-run}, where Dreamer's reconstruction objective struggles to disentangle the task-relevant cheetah dynamics from the moving distractors and plateaus at a substantially lower reward. Momentum Prediction, in contrast, collapses in this setting across all three tasks, staying near its initial performance throughout training. We hypothesize that without an explicit local-global mutual information structure, the prediction target is easily satisfied by encoding distractor motion rather than task-relevant dynamics, whereas our patch-based Deep InfoMax objective encourages the state representation to retain information that is jointly predictive of future local image regions, which distractors do not consistently provide.

\textbf{Natural video setting.} The gap widens further when natural video from Kinetics is used as the background (Figure~\ref{fig:natural}), the most visually challenging of our three settings since the background is neither stationary nor synthetically simple. Here InfoMax substantially outperforms both baselines on every task, where Dreamer's reconstruction loss is dominated by the high-entropy natural video background and the agent fails to make significant task progress. Momentum Prediction likewise plateaus early and fails to escape near-random performance. These results support our central hypothesis: because our objective only requires the state representation to be predictive of future observations, rather than requiring the full observation (background included) to be reconstructed, it is far more robust to visually complex and non-stationary environments that carry less task-relevant signal.

\textbf{Training efficiency.} Beyond final performance, removing the pixel decoder also reduces the per-step compute cost of our method relative to Dreamer, since no pixel decoding network needs to be trained; we found InfoMax to be consistently substantially faster in wall-clock time per training step across all three settings while using equal number of parameters in the encoder and dynamics model.


\section{Conclusion}
We present Contrastive World Models, an approach for training world models without pixel reconstruction. In place of the observation reconstruction term in the RSSM objective, we derive a Deep InfoMax-like lower bound that maximizes the mutual information between state-action sequences and local patch features of future observations. We evaluate our approach on the DeepMind Control Suite across three settings of increasing visual complexity. Our method matches both the reconstruction-based and momentum-based baselines in the simple, distraction-free setting, and significantly outperforms them once simple distractors or natural video backgrounds are introduced. We further observe more efficient training from removing the pixel decoder entirely. Together, these results suggest that Contrastive World Models are a principled and promising direction for building world models that are robust to visual nuisance factors, a property that is particularly relevant for transferring RL agents from simulation to the real world.

\textbf{Limitations:} This work is limited to small-scale experiments built on top of an older architecture (RSSM) on a relatively established benchmark (DeepMind Control Suite). Extending our approach to more recent transformer-based world model architectures, and scaling to large domains, would be a valuable direction for further analysis.

\textbf{Future work:} Our objective is general and makes minimal assumptions beyond access to state-action sequences and future observations, without requiring action-conditioned reconstruction, which makes it straightforward to apply to large-scale, unlabeled pretraining settings as well. Scaling our approach to larger domains, leveraging action-free video data (e.g., from YouTube) for pretraining, and incorporating language-specified goals are all promising directions for future work.


\bibliography{bibliography}

\end{document}